\documentclass[conference,compsoc]{IEEEtran}
\usepackage{multirow}
\usepackage{xcolor}

\ifCLASSOPTIONcompsoc
  \usepackage[nocompress]{cite}
\else
  \usepackage{cite}
\fi
\ifCLASSINFOpdf
   \usepackage[pdftex]{graphicx}
\else
\fi
\usepackage{amsmath}
\begin{document}
%
\title{
A new network-base high-level data classification methodology (Quipus) by modeling attribute-attribute interactions
}


\author{
\IEEEauthorblockN{Esteban Wilfredo Vilca Zuñiga}
\IEEEauthorblockA{Dept. of Computing and Mathematics\\
FFCLRP-USP\\
Ribeirão Preto, Brasil\\
evilcazu@usp.br}
\and
\IEEEauthorblockN{Liang Zhao}
\IEEEauthorblockA{Dept. of Computing and Mathematics\\
FFCLRP-USP\\
Ribeirão Preto, Brasil\\
zhao@usp.br}
}


%


\maketitle

\begin{abstract}

High-level classification algorithms focus on the interactions between instances. These produce a new form to evaluate and classify data. In this process, the core is a complex network building methodology. The current methodologies use variations of kNN to produce these graphs. However, these techniques ignore some hidden patterns between attributes and require normalization to be accurate. In this paper, we propose a new methodology for network building based on attribute-attribute interactions that do not require normalization. The current results show us that this approach improves the accuracy of the high-level classification algorithm based on betweenness centrality.

\end{abstract}


%
\IEEEpeerreviewmaketitle

\section{Introduction}

The machine learning classification algorithms are low level when they use only physical features to classify usually distance measures like euclidean distance and ignore the possible interactions between instances as a system \cite{carneiro2018importanceconcept}. However, high-level classification algorithms focus on exploit theses characteristics using complex networks as data representation \cite{zhao2016mlcn}\cite{carneiro2018GraphsConstruction}.

There are a variety of techniques to build networks but usually, they use kNN as the core \cite{carneiro2018importanceconcept} \cite{seyed2019} \cite{esteban2020bc}. These algorithms produce a network where each node represents an instance and each edge represents a neighbor in kNN \cite{tiago2018hldc}.

A complex network is defined as a non-trivial graph \cite{barabasi2002smcn}. Usually, the quantity of instances on the dataset and the interactions generates a large graph with numerous edges. This large graphs presents special characteristics that are exploited in many techniques to classify data like Betweenness Centrality \cite{esteban2020bc}, Clustering Coefficient \cite{thiago2012hldc}, Assortativity \cite{tiago2018hldc} and so on.

The current methodologies produce one network reducing each instance to a node. This approach presents some different problems like the need to normalize data and the omission of hidden-patterns between attribute-attribute interaction.

In this paper, we will present a new methodology that captures these attribute-attribute interactions building a network for each attribute, removing networks without useful information, optimizing the importance of each one. We will use the high-level classification technique presented in \cite{esteban2020bc} to evaluate the results.

\section{Model Description}
In this section, we describe our algorithm. First, we make a literature review. Then, we describe step by step our algorithm.

\subsection{Literature Review}


\subsubsection{Complex networks as data representation}
\label{subsubsec:complexnetworks}
In order to capture the instance interactions, we need to represent the data in a network structure. 
Different authors present building methodologies using the  k nearest neighbors algorithm.
They transform each instance into a node and the nearest neighbors will be the neighborhood of that node \cite{carneiro2018GraphsConstruction}.
\begin{figure}[h]
    \centering
   	\includegraphics[height=2.0cm]{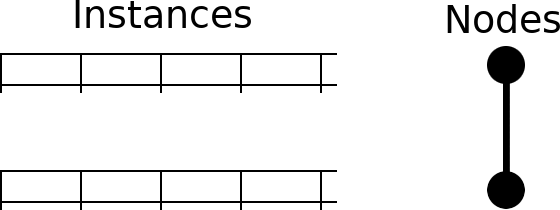}
   	\caption{Image of two related instances transformed in two linked nodes.}
    \label{twoLinked}
\end{figure}

In figure \ref{twoLinked} , we can observe how each instance in the dataset is represented as nodes. The links depend on the k nearest neighbors. Due to this methodology depends on $kNN$, the dataset normalization is needed.

In equation \ref{radius_network_construction_rule}, it is described the general methodology to build the network representation of the data \cite{tiago2018hldc}.

\begin{equation} \label{radius_network_construction_rule}
  \mathcal{N}(X_i)=\begin{cases}
    kNN(X_i,y_i), & \text{otherwise}\\
    \epsilon\text{-}radius(X_i,y_i), & \text{if }|\epsilon\text{-}radius(X_i,y_i)|>k\\
  \end{cases}
\end{equation}

Where $X_i$ is the instance $i$ and $y_i$ is its label. $kNN$ returns the $k$ nearest nodes related with the instance $X_i$ that have the same label $y_i$ using a similarity function like euclidean distance \cite{thiago2012hldc}. 
$\epsilon\text{-}radius$ returns a set of nodes $\{V_j, V_j \in \mathcal{V} :distance(X_i,X_j) < \epsilon \land y_i=y_j \}$. $V_j$ is the node representation of the instance $X_j$. 
The value $\epsilon$ is a percentile of the distances calculated with $kNN$ some authors consider just the median \cite{tiago2018hldc}. 

Following this equation \ref{radius_network_construction_rule}, we build a complex network $\mathcal{G}$ from a dataset $\mathcal{D} = \{ (X_1,y_1),...,(X_n,y_n) \}$. Where each class is represented as sub graph $g^i$.

\begin{figure}[h]
    \centering
   	\includegraphics[height=5.5cm]{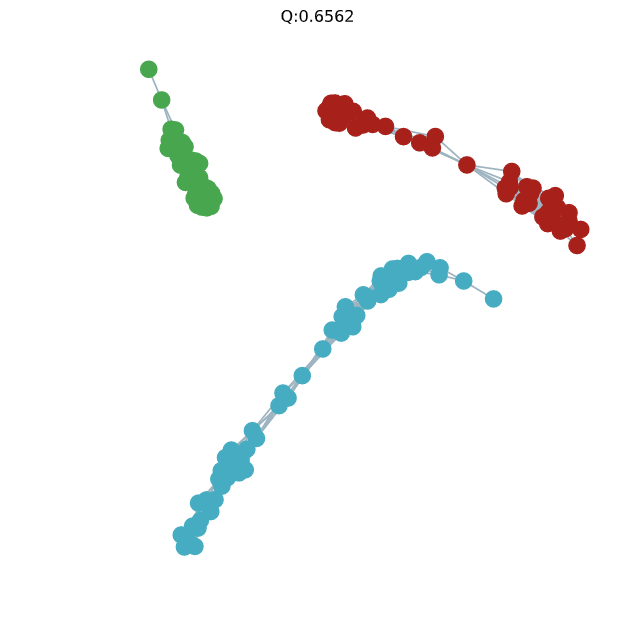}
   	\caption{Image of a complex network $\mathcal{G}$ with three classes $g^{red}$,$g^{green}$,$g^{blue}$ based on Wine UCI dataset and modularity $Q=0.6562$.}
    \label{twoLinked}
\end{figure}
\subsubsection{Important network measures}
To exploit the high-level approach, we need measures that capture nodes interaction like structure, centrality, and connectivity measures.
\begin{itemize}
    \item Clustering Coefficient($CC$): This metric gives information about the connectivity between the neighbors of a node \cite{barabasi2002smcn}. It is between 0 (no communication between the neighbors) and 1 (fully connected neighbors). 
    
    \item Betweenness Centrality ($BC$): This metric capture the communication between nodes using the shortest paths \cite{zhao2016mlcn}. For each node we will calculate the number of geodesic path where this node is present. A node with higher $BC$ present an important role in the network communication.
    \begin{equation} \label{betweenness}
        B(i) = \sum_{s\neq i\in\mathcal{V}} {\sum_{t\neq i\in\mathcal{V}}{ \frac{ \eta^{i}_{st}}{ \eta_{st}}}}
    \end{equation}
    where $\eta^{i}_{st}$ is 1 when the node $i$ is part of the geodesic path from $s$ to $t$ and 0 otherwise. $\eta_{st}$ is the total number of shortest paths between $s$ and $t$.
    
    \item Modularity ($Q$): This metric provides information about the quality of a given partition \cite{Clauset2004}. Usually, it is between 0 and 1. Where 0 means poor community structure and 1 represents a strong differentiation between the communities. In supervised learning classification, the communities are the classes. Higher modularity represents better separations of the sub graphs $g^i$ and probably better classification.
    
    \begin{equation} 
    Q=\frac{1}{2|E|}\sum_{i,j\epsilon V}(A_{ij}-\frac{k_ik_j}{2|E|})
    \label{eq:modularity}
    \end{equation}
    
    Where $A_{ij}$ is the weight of the edge that links the vertex $i$ and $j$. $|E|$ and $V$ represents the number of edges and nodes respectively. $k_i$ is the degree of the node $i$.
    
\end{itemize}
\subsubsection{High-level classification algorithms}
These kinds of algorithms apply these measures in complex networks to classify.
\begin{itemize}
    \item Based on Impact Score (NBHL): This algorithm uses different network measures after and before and insertion\cite{tiago2018hldc}. Where the insertion produces fewer metrics variation the node will be classified.
    \item Based on Importance (PgRkNN): This algorithm use the Pagerank algorithm to evaluate the importance of the nodes \cite{carneiro2018importanceconcept}. For each sub graph $g^i$, we measure the importance of the nodes related to the node to be classified. The neighbors with higher importance will capture the new node.
    \item Based on Betweenness (HLNB-BC): This algorithm insert the new node into each sub graph $g^i$ and search the nodes with similar $BC$ in each sub graph \cite{esteban2020bc}. These are compare and evaluated to provide a probability for each class and the higher is the selected label. 
    
    \begin{equation} \label{eq:hlbc}
        \mathcal{H} \approx (\alpha)\mathcal{W}^n+(1-\alpha)\mathcal{T}^n
    \end{equation}
    
    where $\mathcal{H}$ is a probability list for each label for one node. 
    $\mathcal{W}^n$ is the $BC$ average difference of the $b$ closes nodes. 
    $\mathcal{T}^n$ is the list of the number of link for each subgraph $g^i$. 
    $\alpha$ controls the weights between structural information $\mathcal{W}^n$ and number of links $\mathcal{T}^n$.
\end{itemize}

\subsection{High-Level Classification Algorithm Using Attribute-Attribute Interaction (Quipus)}
\subsubsection{Attribute-attribute interactions}
In the section \ref{subsubsec:complexnetworks}, we analyse how each instance is represented as a node but this approach ignore some hidden patterns between attribute-attribute interaction.

\begin{figure}[h]
    \centering
   	\includegraphics[height=4.5cm]{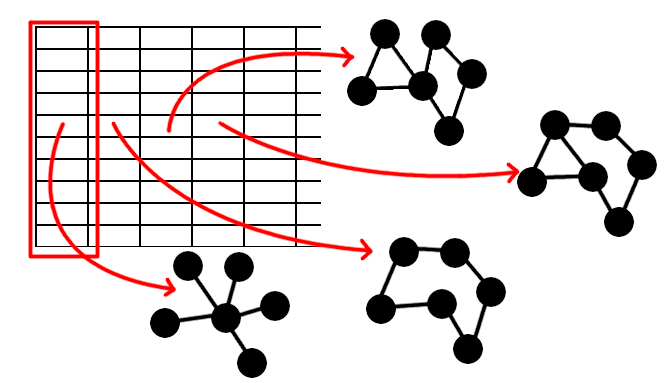}
   	\caption{Image with 4 graphs that capture the interactions into each attribute.}
    \label{fig:4graphs}
\end{figure}

In this paper, we create a graph for each attribute to detect this hidden patterns. In figure \ref{fig:4graphs}, we can appreciate how each attribute is represented as an independent graph. Using this approach, we can capture the attribute-attribute interaction. Since we are using attributes that have the same scale to produces the graphs, our method increases its resistance to non-normalized data.
However, there are attributes that by themselves do not provide relevant information and require others to be useful. Thus, we will use the $Q$ to evaluate each graph.

\subsubsection{Proposed methodology}
\label{subsub:proposed}
In supervised learning, we split the dataset in two $X_{training}$ and $X_{testing}$. In training phase, we produce a model that will help us to predict the testing dataset. In training phase, we need to split the data again in $X_{net}$ and $X_{opt}$ because we have an optimization phase. The proportion depends on the quantity of data available.

The next steps are for training phase:
\begin{itemize}
    \item First, we need to build a graph for each attribute to capture the hidden patters between them following the equation \ref{radius_network_construction_rule} on $X_{net}$. Then, we build one more graph using the instance as a node to capture the instance-instance interaction.
    \item Second, we calculate the $Q$ for each graph. To avoid possible noise of attributes without relevant information, we ignore the graphs with modularity lower than the instance-instance network. 
    \item Third, we insert each instance from $X_{opt}$ to the networks following the same strategy described in step 1. However, we will keep the link between different labels because we want to simulate a real insertion. Introduce each attribute into correspondent graph and the complete instance in the instance-instance graph.
    \item Fourth, we obtain the probability to be part of each class in each graph using the high-level algorithm HLNB-BC. For example, in a dataset with 3 classes and 4 attributes, it will give us a list with three probabilities for each graph (12 probabilities in total).
    \item Fifth, we give a weight for each graph from 0 to 1. This will give us a way to reduce or increase the classification probability of the graphs. Then, we use an optimization algorithm like particle swarm to determinate the better weights for each graph to increase the accuracy of the predicted instances in $X_{opt}$.
    \item Finally, we save the weights and produce the final graphs following the same procedure in step 1 with $X_{training}$.
\end{itemize}
In testing phase, we insert each instance into the graphs following the same process in step 4 of testing phase and multiply the probabilities for each graph with the weights defined in step 5.   



\section{Results} 
In this section, we present the performance of our methodology Quipus. We use Python as programming language, Scikit-learn library for machine learning algorithms \cite{scikit-learn}, networkx as graph library \cite{SciPyProceedings_11}, and Pyswarms as PSO optimizer \cite{pyswarmsJOSS2018}. Each algorithm were tested using 10-folds cross validation 10 times for training and testing datasets, and Grid search to tune the parameters. We search $k$ from 1 to 30, the percentile $\epsilon$ was tested with these values $[0.1,0.2,0.3,0.4,0.5]$, $b$ nearest nodes from 1 to 7 and $\alpha$ with these values $[0.0,0.1,0.2,0.3,0.4,0.5,0.6,0.7,0.8,0.9,1.0]$

The datasets used, their attributes, instances, and classes are described on table \ref{tab:datasets}.
\begin{table}[h!]
    \centering
    \begin{tabular}{ |c|c|c|c|  }
     \hline
     Dataset & Instances  & Attributes & Classes\\
     \hline
     
     Iris &  150 & 4 & 3 \\
     Wine &  178 & 13 & 3 \\
     Zoo &  101 & 16 & 7 \\
     
     \hline
    \end{tabular}
    \caption{information about the uci classification dataset used on these project}
    \label{tab:datasets}
\end{table}

In section \ref{subsub:proposed}, we split the training data in $X_{net}$ and $X_{opt}$. In our tests, we use a stratified random split 80\% and 20\% respectively. This value could be modified according to the quantity of data. 

Then, we build a network for each attribute and one network for instance-instance interactions. We calculate their modularities ($Q$) and compare each attribute network with the instance-instance network. The networks with lower modularity will be ignored in the rest of the process. The table \ref{tab:modularities} show us the modularities for each network.

\begin{table}[h]
    \centering
    \begin{tabular}{|c|c|}
        \hline
        Network & Modularity $Q$ \\
        \hline
         Instance-instance &  0.3181 \\
         \hline
         Attribute 1 &  0.3189 \\
         \hline
         Attribute 2 &  0.0924 \\
         \hline
         Attribute 3 &  0.0500 \\
         \hline
         Attribute 4 &  0.1689 \\
         $\vdots$ & $\vdots$\\
         Attribute 10 &  0.2288 \\
         \hline
         Attribute 11 &  0.3008 \\
         \hline
         Attribute 12 &  0.3333 \\
         \hline
    \end{tabular}
    \caption{Modularities of attribute networks in UCI Wine Dataset}
    \label{tab:modularities}
\end{table}

For instance, the modularity of the networks attribute 1 and attribute 12 are higher than modularity of instance-instance network. So, these networks will be used for optimization, classification, and insertion. The others will be ignore because do not have a high community structure.

The insertion of the nodes into each graph follow the equation \ref{radius_network_construction_rule}, but preserving the links with nodes of different labels. Given that we want to capture the insertion probability for each class. Then, we create a weight for each graph probabilities and start an optimization phase. We use a particle swarm optimization from Pyswarms library with these parameters $\{c_1=0.5,c_2=0.1,w=0.9, iterations=500\}$. Theses could be optimized but we use these fixed values for these experiments.

For example, in one interaction, our algorithm capture the weight in table \ref{tab:modularities}.
\begin{table}[h]
    \centering
    \begin{tabular}{|c|c|c|c|}
        \hline
        Network & Modularity $Q$ & Weights & Ignored\\
        \hline
         Instance-instance &  0.3181 & 0.9083 & False \\
         \hline
         Attribute 1 &  0.3189 & 0.8065 & False\\
         \hline
         Attribute 2 &  0.0924 & - & True \\
         \hline
         Attribute 3 &  0.0500 & - & True \\
         \hline
         Attribute 4 &  0.1689 & - & True \\
         $\vdots$ & $\vdots$\\
         Attribute 10 &  0.2288 & - & True \\
         \hline
         Attribute 11 &  0.3008 & - & True \\
         \hline
         Attribute 12 &  0.3333 & 0.1746 & False \\
         \hline
    \end{tabular}
    \caption{Modularities and weights of attribute networks in UCI Wine Dataset}
    \label{tab:modularities}
\end{table}

Once the weights are defined, we proceed to rebuild the graphs but using the entire training dataset $X_{training}$. Finally, the classification phase, will follow the same process that optimization phase but using the optimized weights.
\begin{figure}[h!]
    \centering
   	\includegraphics[height=6cm]{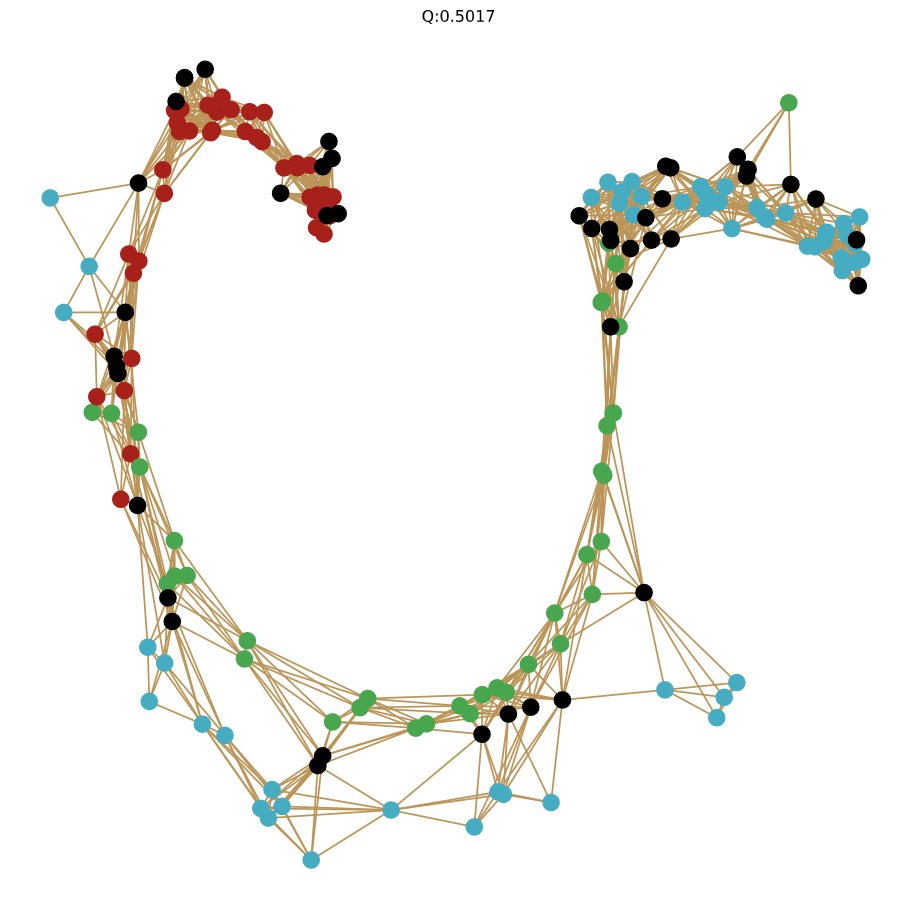}
   	\caption{Image of instance-instance network from one interaction in UCI wine dataset classification.}
    \label{fig:iinetowrk}
\end{figure}
\begin{figure}[h!]
    \centering
   	\includegraphics[height=6cm]{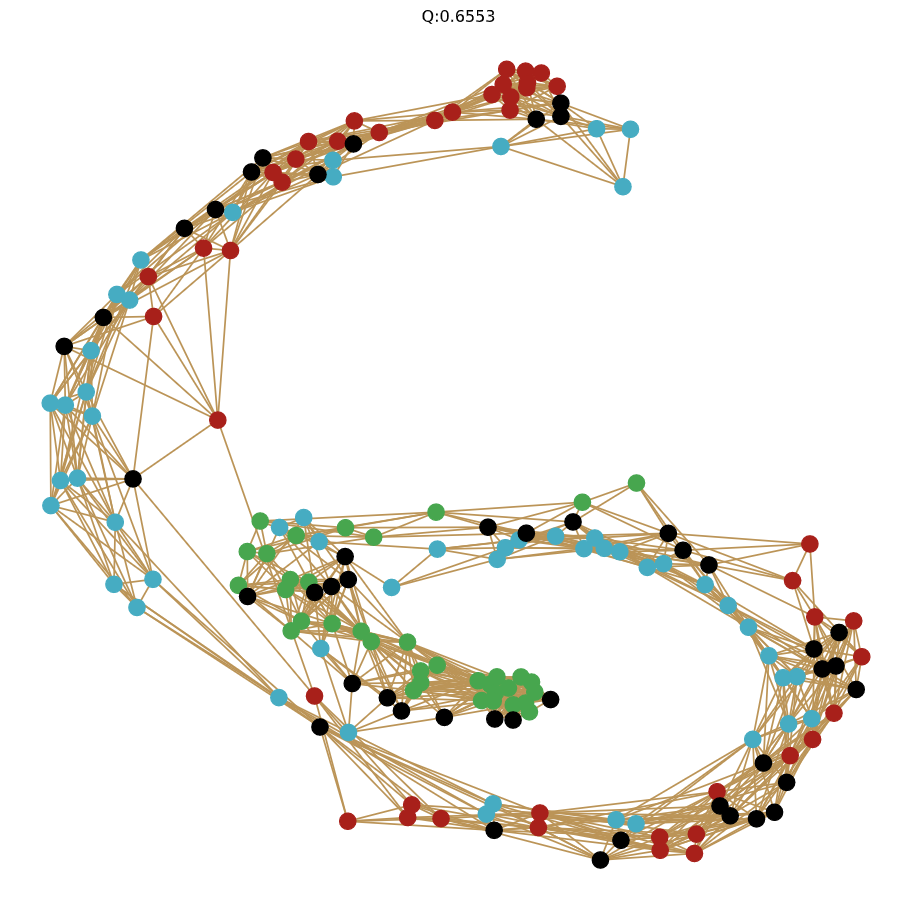}
   	\caption{Image of attribute-attribute 1 network from one interaction with higher modularity than its instance-instance network in UCI wine dataset classification.}
    \label{fig:aanetwork}
\end{figure}

In figure \ref{fig:iinetowrk} , we can observe the instance-instance network from one instance of wine dataset classification. The black nodes represents instances classified. It present an structure where the red nodes are in one side, the blue nodes in the other side, and the green nodes in the middle of them. In the figure \ref{fig:aanetwork}, we observe the network from the first attribute of wine datset that had a modularity of 0.3189. Once the nodes are inserted this graphs present a higher modularity $Q=0.6553$. Without this methodology, we will lose these attribute-attribute interactions. These networks gives us an accuracy of 91.11\%.

\begin{table}[h!]
    \centering
    \begin{tabular}{ |c|c|c|c|c|  }
     \hline
     Dataset & $k$ & $e$ & $b$ & $\alpha$ \\
     \hline
     
     Iris  &    12  & 0.0 & 3 & 1.0 \\
     Wine &     7  & 0.0 & 3 & 1.0 \\
     Zoo &      1   & 0.0 & 1 & 1.0 \\
     
     \hline
    \end{tabular}
    \caption{parameter values used by nbhl-bc with Quipus methodology in uci datasets}
    \label{tab:parametersDataset}
\end{table}

In table \ref{table:resultAccuracy}, we observe the accuracy of Quipus against the literature network building technique kNN+$\epsilon$-$radius$. This current technique present problems related to data non-normalized like wine uci dataset. However, using Quipus, we reduce this problem. Due to the attribute networks build their relations in the same scale, the optimized weight manage the probability force, and reduce its impact in the final classification.

\begin{table}[h!]
\centering
\begin{tabular}{ |c|c|c|c| } 
 \hline
 \multicolumn{4}{|c|}{Results of 10 times using 10-folds cross validation} \\
\hline
Dataset & Prediction  & Building (k) & Accuracy \\

\hline
\multirow{2}{4em}{Iris} & \multirow{2}{5em}{HLNB-BC} & 
kNN+$\epsilon$-$radius$ (7)  & 95.33 $\pm$ 11.85 \\ 
& &
Quipus (12)  & 95.80 $\pm$ 09.36 \\ 
\hline
\multirow{2}{4em}{Wine} & \multirow{2}{5em}{HLNB-BC} & 
kNN+$\epsilon$-$radius$ (1)  & 75.84 $\pm$ 19.15 \\ 
& &
Quipus (7)  & 93.03 $\pm$ 13.08 \\ 
\hline
\multirow{2}{4em}{Zoo} & \multirow{2}{5em}{HLNB-BC} & 
kNN+$\epsilon$-$radius$ (1)  & 96.36 $\pm$ 12.98 \\ 
& &
Quipus(1)  & 96.87 $\pm$ 04.97 \\ 
\hline
\end{tabular}
\caption{Table with accuracy of different building methodologies and UCI datasets without normalization. }
\label{table:resultAccuracy}
\end{table}

\section{Conclusion}
The new classification methodology proposed exploit the hidden patterns in attribute-attribute interactions. Building networks for each attribute and ignoring the ones with lower modularity. Also, uses the high-level classification technique HLNB-BC. Introduces resilience to the model against non-normalized data.

Many different modification, tests, and experiments have been left for future work like testing with others high-level techniques (NBHL, PgRkNN), identify a way to optimize the parameters of particle swarm, and the use of the Quipus methodology in other real datasets.






\bibliographystyle{IEEEtran}

%
\bibliography{bibliography.bib}




\end{document}